\documentclass[]{bytedance_seed}
\usepackage{algorithm}
\usepackage{algorithmic}
\usepackage{amsmath}
\usepackage{amssymb}
\usepackage{makecell}
\usepackage{colortbl}
\microtypesetup{expansion=false}
\hypersetup{
  pdftitle={Training-Free Predictive Spatio-Temporal Token Pruning for Few-Step Autoregressive Video Diffusion},
  pdfauthor={Jia Li, Xurui Peng, Xing Wang, Hayden Kwok-Hay So}
}

\definecolor{oursbg}{RGB}{234,240,248}
\definecolor{dltcolor}{RGB}{90,110,140}
\newcommand{\hl}{\cellcolor{oursbg}}
\newcommand{\dlt}[1]{\raisebox{0.15ex}{\tiny\textcolor{dltcolor}{$#1$}}}
\newcommand{\total}[2]{#1\,\dlt{#2}}
\definecolor{gainclr}{RGB}{200,30,30}
\newcommand{\gain}[1]{\raisebox{0.15ex}{\tiny\textcolor{gainclr}{$#1$}}}
\newcommand{\totalgain}[2]{#1\,\gain{#2}}

\title{Training-Free Predictive Spatio-Temporal Token Pruning for Few-Step Autoregressive Video Diffusion}
\author[1,2,*]{Jia Li}
\author[2,*]{Xurui Peng}
\author[2,\dagger]{Xing Wang}
\author[1,\dagger]{Hayden Kwok-Hay So}

\affiliation[1]{The University of Hong Kong}
\affiliation[2]{ByteDance Seed}

\contribution[*]{Equal contribution}
\contribution[\dagger]{Corresponding authors}

\abstract{
Few-step autoregressive (AR) video diffusion models generate videos chunk by chunk and have made efficient video generation practical. However, further accelerating these models toward real-time generation remains challenging, as few-step distillation leaves little cross-step redundancy for existing training-free caching methods to exploit. We revisit this regime and identify two token-level redundancy patterns: cross-frame redundancy from static content shared across adjacent frames, and cross-step redundancy from tokens that converge early during denoising. Crucially, we find that early clean estimates along the model's own denoising trajectory provide an effective online signal for locating both patterns, requiring no retraining, auxiliary predictor, or reference video. Based on these findings, we propose \textbf{PreSTP}
(\textbf{Pre}dictive \textbf{S}patio-\textbf{T}emporal Token
\textbf{P}runing), a training-free framework that constructs a unified
token-wise redundancy mask from cross-frame similarity and cross-step convergence. Tokens marked as redundant skip subsequent computation under a consistent last-computed-value restoration rule, and a content-adaptive mask reuse strategy further reduces dense probing cost. On 5-second generation, PreSTP achieves $1.52\times$--$3.20\times$ speedups and a better efficiency--fidelity trade-off than coarse-grained caching baselines such as TeaCache and FlowCache. On 30-second generation, it reaches up to $1.96\times$ speedup while improving VBench-Long and suppressing long-horizon drift. Notably, it continues to reduce inference cost even in the extreme one-step regime, where cross-step redundancy vanishes entirely and coarse step-wise caching is no longer applicable.
}

\begin{document}

\maketitle

\section{Introduction}

Diffusion models have become the dominant approach for high-quality video generation, but their iterative denoising process remains computationally expensive. Recent autoregressive (AR) video diffusion models generate videos chunk by chunk with causal attention and have emerged as a promising paradigm for interactive and long-duration synthesis~\cite{teng2025magi,chen2025skyreels}. These early AR models, however, still require tens of denoising steps per chunk, resulting in substantial generation cost. Few-step models such as CausVid and Self-Forcing reduce per-chunk denoising to only four steps~\cite{yin2025slow,huang2026self}, making efficient AR video generation practical.

However, further reducing their latency toward real-time generation is difficult. Most training-free acceleration approaches reuse computation between neighboring denoising steps: TeaCache and related methods cache intermediate features across steps~\cite{liu2025timestep,lyu2025fastercache}, while FlowCache applies step-wise feature reuse to AR video generation~\cite{ma2026flow}. Although effective for models with longer sampling schedules, these methods have much less step-axis redundancy to exploit after few-step distillation. In this regime, aggressive step reuse can noticeably perturb the generation trajectory while providing limited speedup.

We analyze few-step AR video diffusion and identify two forms of token-level redundancy: (i) \emph{cross-frame redundancy}, where spatially aligned tokens remain similar across adjacent frames because of shared static content, and (ii) \emph{cross-step redundancy}, where different tokens converge at different stages of denoising. Moreover, although the clean video is unavailable during text-to-video generation, early clean estimates along the model's own denoising trajectory provide a direct online signal for identifying both types of redundancy, without an external predictor or reference video.

Based on these observations, we propose \textbf{PreSTP}, a training-free framework for predictive spatio-temporal token pruning in few-step AR video diffusion. For each chunk, we use the model's own clean estimates to construct a unified token-wise mask combining cross-frame redundancy with cross-step convergence. Unlike coarse step-wise caching, PreSTP operates online at token granularity. Redundant tokens skip subsequent transformer computation and reuse their most recent values, while temporal memory remains consistent. A content-adaptive mechanism reuses masks across chunks to reduce dense probing overhead for static or slowly changing content. The method requires no retraining, auxiliary predictor, or external reference video.

Extensive experiments demonstrate consistent gains across diverse few-step AR video diffusion models. On short-video generation, our method accelerates CausVid, Self-Forcing, and AnyFlow by $1.52\times$--$3.20\times$ while keeping the VBench score within a small margin of the original model, clearly outperforming TeaCache and FlowCache on the speed-fidelity trade-off. On 30-second generation, it reaches up to $1.96\times$ speedup and, notably, even improves the VBench-Long score over the dense baseline by suppressing long-horizon drift. The same principle generalizes to the extreme one-step regime, where previous step-wise caching methods are no longer applicable. On One-Forcing and Causal Forcing++, it attains $1.10\times$--$1.22\times$ speedups with well-preserved generation quality, and even surpasses dense models in VBench-Long score under the 30-second setting.

Our contributions are summarized as follows:
\begin{itemize}
    \item We identify two token-level redundancy patterns in few-step AR video diffusion: cross-frame static redundancy and cross-step convergence redundancy. We show that early clean estimates during denoising provide an effective online signal for identifying both patterns without requiring an auxiliary predictor or external reference video.

    \item We propose PreSTP, a training-free framework that unifies both redundancy patterns into an online token-wise mask and skips redundant token computation. Last-computed-value restoration preserves output and KV-cache consistency, while content-adaptive mask reuse reduces dense probing to further accelerate inference.

    \item We validate our method across a wide range of few-step and one-step AR video diffusion models, achieving up to $3.20\times$ speedup with negligible loss in visual quality. It provides a stronger speed--fidelity trade-off than training-free caching baselines and remains effective in the one-step regime where such baselines no longer apply.
\end{itemize}

\section{Related Work}

\paragraph{Few-Step Autoregressive Video Diffusion.}
Recent autoregressive (AR) video diffusion models adopt causal attention and
generate videos chunk by chunk. Representative models such as CausVid and
Self-Forcing use distillation to accelerate generation~\cite{
yin2025slow,huang2026self}, while subsequent works target longer durations~\cite{yang2025longlive,li2026train} or fewer denoising network evaluations per chunk~\cite{
feng2026oneforcingstableonestepautoregressive,
zhao2026causalforcingscalablefewstep,li2026tritd}. In these distilled models, denoising-trajectory redundancy is already substantially reduced, limiting the effectiveness of conventional step-wise acceleration.

\paragraph{Training-Free Diffusion Acceleration.}
Most training-free acceleration methods target multi-step diffusion models and reduce computation along the denoising trajectory.
TCFG truncates redundant text-guidance computation~\cite{Fu_2025_ICCV}, while
TeaCache and related methods reuse intermediate features across steps~\cite{
liu2025timestep,lyu2025fastercache}. FlowCache further introduces
chunk-adaptive feature caching and KV compression for AR video~\cite{
ma2026flow}. However, few-step models offer few adjacent states for reuse, sharply limiting the available step-wise redundancy.
Other approaches instead reduce the cost of historical attention through KV compression or instance-specific parametric absorption~\cite{ji2026forcing,fu2026towards}.

\paragraph{Token Reduction and Pruning.}
Existing methods reduce diffusion computation by dynamically scheduling text-conditioning tokens~\cite{Li_2026_CVPR} or combining sparse attention with visual-token budget allocation~\cite{zhang2026training,liu2025astraea}. Beyond reducing computation within each step, recent training-free methods vary token update frequencies according to motion or denoising dynamics~\cite{xu2026motion,chu2026not}. In image and video editing, reference inputs provide an additional signal for identifying unedited regions or duplicated latent patches, allowing redundant computation to be skipped~\cite{chen2025regione,menn2026video}. Token pruning also serves a different purpose in AR long-video generation: TokenTrim removes unstable context tokens and regenerates the current chunk to suppress temporal drift, introducing additional inference cost~\cite{shaulov2026tokentrim}. In contrast, PreSTP targets acceleration of few-step AR generation from noise. It identifies cross-frame and cross-step redundancy online using clean estimates along the model's own denoising trajectory and skips redundant token computation, without requiring an external reference image or video, retraining, or an auxiliary prediction model.

\section{Method}
\subsection{Preliminaries}

Autoregressive video diffusion models typically generate in a chunk-wise or frame-wise manner.
For simplicity of notation, we take frame-wise generation as the main example, where a video latent is partitioned into $N$ temporal frames:
\begin{equation}
    X = \{x^{(1)}, x^{(2)}, \ldots, x^{(N)}\}.
\end{equation}
Given a condition $c$ (e.g., a text prompt), the autoregressive factorization is:
\begin{equation}
    p_{\theta}(X \mid c) = \prod_{i=1}^{N} p_{\theta}\left(x^{(i)} \mid x^{(<i)}, c\right),
\end{equation}
where $x^{(<i)}$ denotes previously generated historical frames, summarized via the autoregressive KV cache $\mathcal{K}^{(<i)}$.

For the $i$-th frame, generation starts from pure noise and runs $K$ denoising steps with $t_0 > t_1 > \cdots > t_K = 0 $. Under the flow-matching parameterization, the denoising network $\Phi_{\theta}$ predicts a velocity field
\begin{equation}
    v_{\theta}^{(i,k)} \triangleq \Phi_{\theta}\!\left(x_{t_k}^{(i)}, t_k, c, \mathcal{K}^{(<i)}\right),
\end{equation}
conditioned on the current latent, timestep $t_k$, text prompt $c$, and historical KV cache $\mathcal{K}^{(<i)}$. The solver uses $v_{\theta}^{(i,k)}$ to advance the latent,
\begin{equation}
    x_{t_{k+1}}^{(i)} = x_{t_k}^{(i)} + (t_{k+1}-t_k)\,v_{\theta}^{(i,k)},
    \quad k=0,\ldots,K-1,
\end{equation}
and it also yields the clean latent estimate at step $k$,
\begin{equation}
    \hat{x}_{0\mid t_k}^{(i)}
    =
    x_{t_k}^{(i)} - t_k\,v_{\theta}^{(i,k)}.
    \label{eq:x0est}
\end{equation}

The $i$-th latent frame consists of a set of spatial tokens $\mathcal{I}^{(i)}$, where each token $j \in \mathcal{I}^{(i)}$ corresponds to a specific spatial patch. In the standard dense generator, all tokens in $\mathcal{I}^{(i)}$ are recomputed at every denoising step, regardless of whether they have already stabilized.

\begin{figure}[t!]
    \centering
    \includegraphics[width=0.80\textwidth]{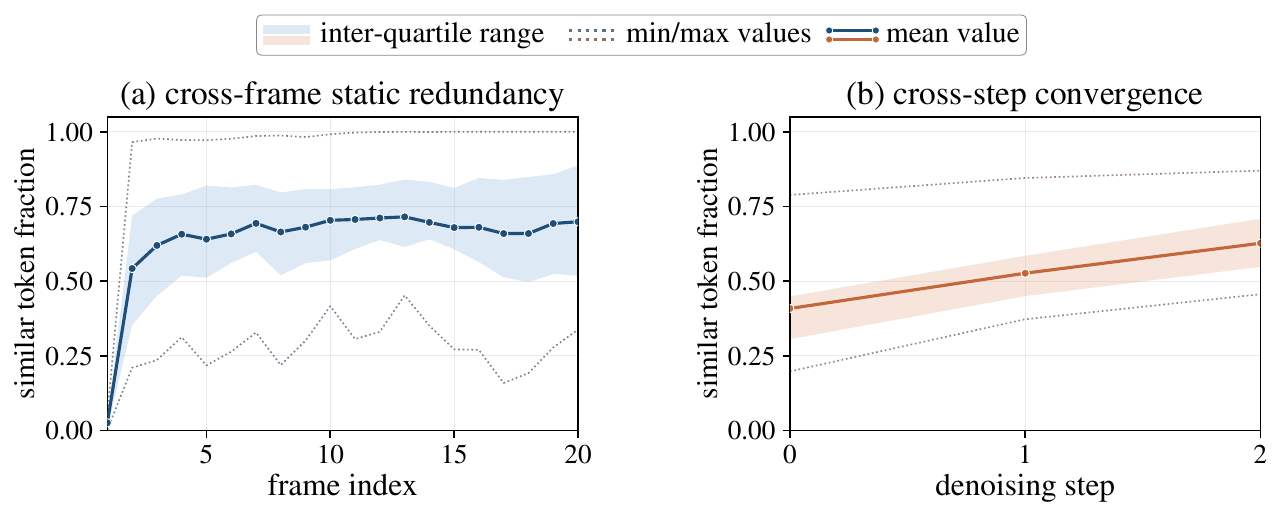}
    \vspace{-4mm}
    \caption{Token-level redundancy analysis on Self-Forcing.
    }
    \label{fig:observation}
    \vspace{-5mm}
\end{figure}

\subsection{Observation: Heterogeneous Token Redundancy}
\label{sec:observation}

We measure token stability by the cosine similarity between two of its clean latents or estimates, and treat a token as stable when the similarity exceeds a threshold. Under this criterion, we identify two types of redundancy in AR video diffusion models, summarized in
Figure~\ref{fig:observation} ($\tau_s=\text{0.95}$,
$\tau_f=\text{0.99}$).

\emph{(1) Cross-frame spatial redundancy.}
Many tokens remain nearly unchanged between adjacent frames,
\begin{equation}
    \cos\!\big([x_{0}^{(i)}]_j,
           [x_{0}^{(i-1)}]_j\big) > \tau_s,
\end{equation}
especially in static or slowly changing background regions. Such tokens are largely determined by the historical context and gain little from repeated recomputation.

\emph{(2) Cross-step temporal redundancy.}
During denoising, many tokens yield highly similar clean predictions across consecutive steps,
\begin{equation}
    \cos\!\big([\hat{x}_{0\mid t_k}^{(i)}]_j, [\hat{x}_{0\mid t_{k+1}}^{(i)}]_j\big) > \tau_f,
\end{equation}
indicating that these tokens have stabilized across steps.

Crucially, this stability is not uniform across tokens, and it emerges early in the trajectory: a substantial fraction of tokens are largely settled after the very first denoising step. 

\begin{figure*}[t!]
    \centering
    \includegraphics[width=0.9\textwidth]{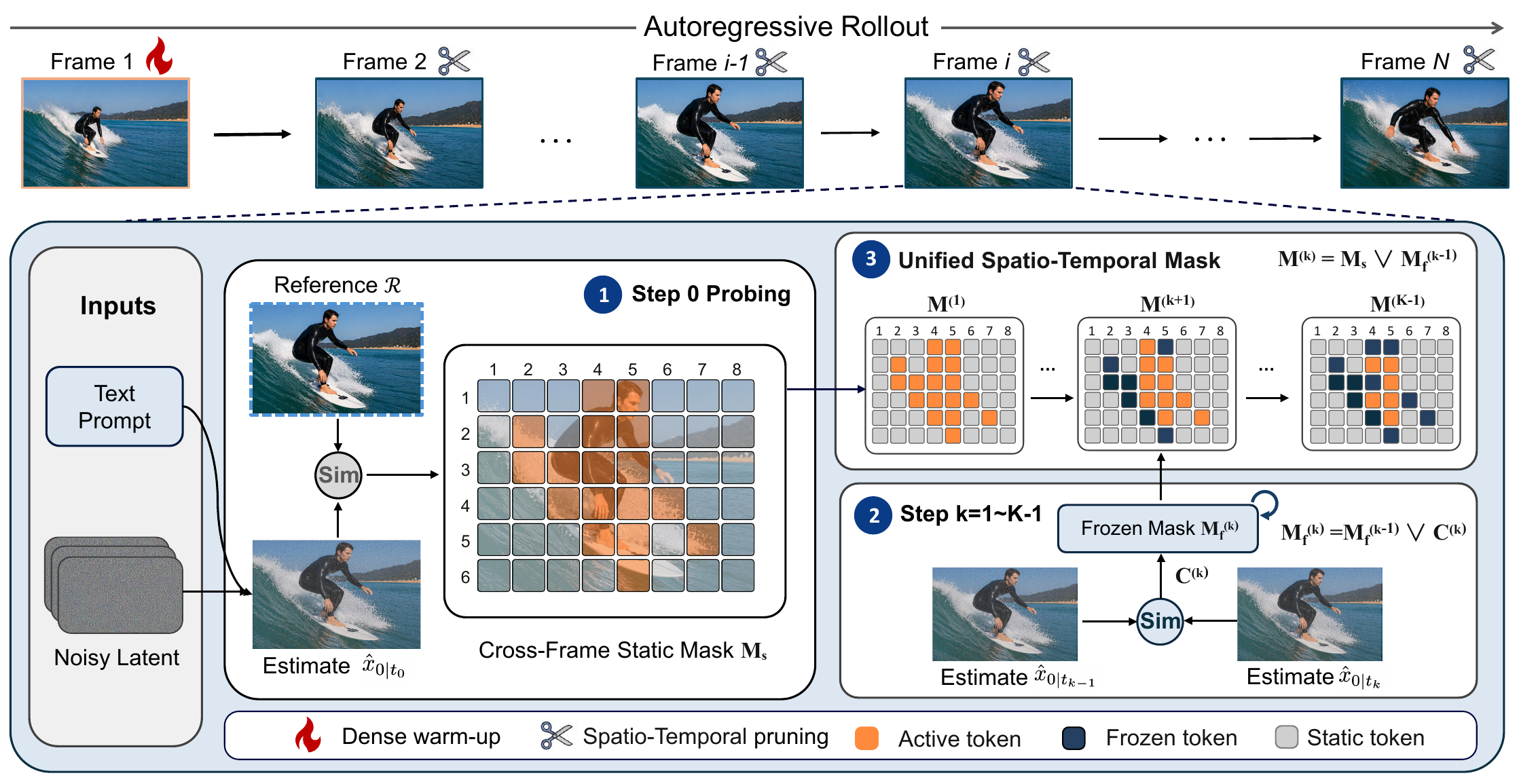}
    \caption{Overview of PreSTP. After the dense first-frame warm-up,
the step-0 clean estimate is compared with the previous-frame
reference to predict $M_s$. At each later denoising step,
consecutive clean estimates identify newly converged tokens $C$
and update $M_f$. Together, $M_s$ and $M_f$ form the next-step
mask $M$, which selects active tokens for computation.}
    \label{fig:framework}
\vspace{-4mm}
\end{figure*}

\begin{algorithm}[t]
\caption{Predictive Spatio-Temporal Token Pruning}
\label{alg:pstp}
\begin{algorithmic}[1]
\REQUIRE Initial latents $\{x_{t_0}^{(i)}\}_{i=1}^{N}$,
condition $c$, timesteps $\{t_k\}_{k=0}^{K}$,
thresholds $\tau_s,\tau_f,g$, re-anchor period $\Delta$

\STATE Generate frame $1$ densely; set $\rho_1\leftarrow0$ and initialize $R^{(2)}$ and $\mathcal{K}^{(<2)}$
\FOR{$i=2$ to $N$}
    \STATE $M_f^{(i,0)} \leftarrow \mathbf{0}$

    \IF{$i=2$ or $\rho_{i-1}<g$ or $(i-2)\bmod\Delta=0$}
        \STATE $v_\theta^{(i,0)},\hat{x}_{0\mid t_0}^{(i)}
        \leftarrow
        \Phi_\theta(x_{t_0}^{(i)},t_0,c,\mathcal{K}^{(<i)})$

        \STATE $M_s^{(i)}(j)\leftarrow
        \mathbb{I}\!\left[
        \cos\!\left(
        [\hat{x}_{0\mid t_0}^{(i)}]_j,
        [R^{(i)}]_j
        \right)>\tau_s
        \right]$
    \ELSE
        \STATE $M_s^{(i)}\leftarrow M_s^{(i-1)}$

        \STATE $S^{(i,0)}\leftarrow
        \{j\in\mathcal{I}^{(i)}
        \mid M_s^{(i)}(j)=0\}$

        \STATE $v_\theta^{(i,0)},\hat{x}_{0\mid t_0}^{(i)}
        \leftarrow
        \Phi_\theta(x_{t_0}^{(i)},t_0,c,\mathcal{K}^{(<i)};
        S^{(i,0)})$
    \ENDIF

    \STATE $x_{t_1}^{(i)}\leftarrow
    x_{t_0}^{(i)}+(t_1-t_0)v_\theta^{(i,0)}$

    \FOR{$k=1$ to $K-1$}
        \STATE $M^{(i,k)}
        \leftarrow M_s^{(i)}\vee M_f^{(i,k-1)}$

        \STATE $S^{(i,k)}\leftarrow
        \{j\in\mathcal{I}^{(i)}
        \mid M^{(i,k)}(j)=0\}$

        \STATE $v_\theta^{(i,k)},\hat{x}_{0\mid t_k}^{(i)}
        \leftarrow
        \Phi_\theta(x_{t_k}^{(i)},t_k,c,\mathcal{K}^{(<i)};
        S^{(i,k)})$

        \STATE $C^{(i,k)}(j)\leftarrow
        \mathbb{I}\!\left[
        \cos\!\left(
        [\hat{x}_{0\mid t_k}^{(i)}]_j,
        [\hat{x}_{0\mid t_{k-1}}^{(i)}]_j
        \right)>\tau_f
        \right],\quad j\in S^{(i,k)}$

        \STATE $M_f^{(i,k)}
        \leftarrow M_f^{(i,k-1)}\vee C^{(i,k)}$

        \STATE $x_{t_{k+1}}^{(i)}\leftarrow
        x_{t_k}^{(i)}
        +(t_{k+1}-t_k)v_\theta^{(i,k)}$
    \ENDFOR

    \STATE $\rho_i\leftarrow\frac{|M_s^{(i)}|}{|\mathcal{I}^{(i)}|}$

    \STATE $R^{(i+1)},\mathcal{K}^{(<i+1)}
    \leftarrow\mathrm{UpdateMemory}(i)$
\ENDFOR
\end{algorithmic}
\end{algorithm}

\subsection{Online Spatio-Temporal Masking}
\label{sec:online_masking}
Motivated by the heterogeneous, early-emerging redundancy observed above, we
propose the online pruning pipeline illustrated in
Figure~\ref{fig:framework} and Algorithm~\ref{alg:pstp}, which replaces dense execution on the full
token set $\mathcal{I}^{(i)}$ with a dynamic active set
\begin{equation}
    S^{(i,k)} \subseteq \mathcal{I}^{(i)},
\end{equation}
so that only tokens in $S^{(i,k)}$ are explicitly computed at timestep $k$, while other tokens are masked and skipped.

The main difficulty is to determine $S^{(i,k)}$ online. Video-editing methods can derive a reuse pattern in advance from a given reference video~\cite{menn2026video}, but no such reference exists during autoregressive generation, so the pruning mask must be predicted from the model's own trajectory. 

\paragraph{Cross-frame static mask.} 
For frame $i$, reference $R^{(i)}$ is derived from frame $i-1$. Each token in $R^{(i)}$ stores its \emph{last-computed clean
value}: the final clean latent if it completes the denoising
schedule, or the clean estimate from its last executed step if
frozen early. Thus, $R^{(i)}$ provides a consistent per-token
reference regardless of how many denoising steps each token
executes. For each token $j \in \mathcal{I}^{(i)}$, we compare its initial clean estimate with the corresponding reference token:
\begin{equation}
    M_s^{(i)}(j)
    =
    \mathbb{I}
    \!\left[
    \cos\!\left(
        [\hat{x}_{0\mid t_0}^{(i)}]_j,
        [R^{(i)}]_j
    \right)
    >
    \tau_s
    \right].
\end{equation}
Here, $\hat{x}_{0\mid t_0}^{(i)}$ denotes the clean estimate from the first denoising step. In few-step generators, this early estimate is informative enough to serve as a proxy for identifying cross-frame static tokens. $M_s^{(i)}(j)=1$ indicates that token $j$ can be skipped in subsequent computation. Once constructed, $M_s^{(i)}$ remains fixed within the current frame.

\paragraph{Cross-step frozen mask.}
The static mask captures inter-frame redundancy relative to the historical reference, but some non-static tokens also converge before the denoising schedule completes, exhibiting intra-frame redundancy. To exploit this, we maintain a growing frozen mask $M_f^{(i,k)}$ that records tokens whose clean estimates have stabilized across denoising steps. Since a skipped token produces no new estimate, we evaluate convergence only for tokens that remain active at step $k$, i.e., those in the keep set $S^{(i,k)}$. For each such token, we compare its clean estimate at step $k$ with that at step $k-1$, and indicate whether it has newly converged:
\begin{equation}
    C^{(i,k)}(j)
    =
    \mathbb{I}\!\left[
        \cos\!\left(
            [\hat{x}_{0\mid t_k}^{(i)}]_j,\,
            [\hat{x}_{0\mid t_{k-1}}^{(i)}]_j
        \right) > \tau_f
    \right],
    \qquad j\in S^{(i,k)},
\end{equation}
where $\tau_f$ is the freezing threshold and $C^{(i,k)}(j)=1$ marks a token that has newly converged at step $k$; we set $C^{(i,k)}(j)=0$ for $j\notin S^{(i,k)}$. The frozen mask is then the running accumulation of these per-step increments,
\begin{equation}
M_f^{(i,k)}
=
M_f^{(i,k-1)} \vee C^{(i,k)},
\qquad M_f^{(i,0)}=\mathbf{0},
\end{equation}
so that once a token is marked as converged it remains frozen for all remaining denoising steps of the same frame.

\paragraph{Unified skip mask and keep set.}
At denoising step $k$, we combine the two heterogeneous masks into a unified mask that merges cross-frame redundancy with previously detected cross-step convergence:
\begin{equation}
    M^{(i,k)}
    =
    M_s^{(i)} \vee M_f^{(i,k-1)},
\end{equation}
and define the keep set as the tokens not marked by $M^{(i,k)}$,
\begin{equation}
    S^{(i,k)}
    =
    \left\{
    j \in \mathcal{I}^{(i)}
    \mid
    M^{(i,k)}(j)=0
    \right\}.
\end{equation}
Only the tokens in $S^{(i,k)}$ are executed by the transformer at step $k$; the tokens marked in $M^{(i,k)}$ are skipped and restored according to the rules introduced in the following subsection.

\paragraph{Restoration and KV-cache consistency.}
We use $\Phi_{\theta}(\cdot;S)$ to denote evaluation on the active set $S$
followed by restoration to the full token layout. Skipped token outputs and
KV-cache entries retain their most recently computed values, while only active
cache entries are updated. Further details are provided in the supplementary
material.

\paragraph{First-frame dense warm-up.}
In autoregressive video generation, the first frame serves as the reference anchor for subsequent frames, so early approximation errors can propagate throughout the rollout. Moreover, the cross-frame static mask is defined by the cosine similarity between each frame and its predecessor, and it cannot be computed for the first frame, which has no preceding frame for comparison. We therefore generate the first frame with the original dense computation and restrict predictive token pruning to the subsequent frames. This design naturally handles the boundary at the sequence start and mitigates initial error accumulation, and it aligns with the first-frame enhancement (FFE) principle in ASD~\cite{yang2025towards}, which highlights the role of the initial frame in stabilizing long autoregressive generation.

\begin{table*}[!t]
\centering
\small
\setlength{\tabcolsep}{5pt}
\resizebox{\textwidth}{!}{%
\begin{tabular}{ll ccc ccc ccc}
\toprule
\multirow{2}{*}{Model} & \multirow{2}{*}{Method}
 & \multirow{2}{*}{Latency(s)$\downarrow$}
 & \multirow{2}{*}{Speedup$\uparrow$}
 & \multirow{2}{*}{\makecell{Throughput\\(FPS)$\uparrow$}}
 & \multirow{2}{*}{LPIPS$\downarrow$}
 & \multirow{2}{*}{PSNR$\uparrow$}
 & \multirow{2}{*}{SSIM$\uparrow$}
 & \multicolumn{3}{c}{VBench$\uparrow$} \\
\cmidrule(lr){9-11}
 & & & & & & & & Quality & Semantic & Total \\
\midrule
\multirow{4}{*}{CausVid}
 & Vanilla   & 7.86 & 1.00$\times$ & 10.3 & --    & --    & --    & 84.02 & 78.46 & 82.91 \\
 & +TeaCache  & 6.30 & 1.25$\times$ & 12.9 & 0.286 & 15.02 & 0.784 & 83.79 & 78.09 & \total{82.65}{-0.26} \\
 & +FlowCache & 4.97 & 1.58$\times$ & 16.3 & 0.456 & 12.28 & 0.608 & 83.73 & \textbf{79.36} & \total{\textbf{82.86}}{-0.05} \\
 & \hl\textbf{+PreSTP} & \hl\textbf{4.65} & \hl\textbf{1.69$\times$} & \hl\textbf{17.4}
   & \hl\textbf{0.092} & \hl\textbf{21.03} & \hl\textbf{0.934}
   & \hl\textbf{83.89} & \hl 78.69 & \hl\total{82.85}{-0.06} \\
\midrule
\multirow{4}{*}{Self-Forcing}
 & Vanilla   & 7.92 & 1.00$\times$ & 10.2 & --    & --    & --    & 84.67 & 81.26 & 83.99 \\
 & +TeaCache  & 6.31 & 1.26$\times$ & 12.8 & 0.430 & 14.19 & 0.691 & 84.59 & 80.35 & \total{83.74}{-0.25} \\
 & +FlowCache & 6.31 & 1.26$\times$ & 12.8 & 0.428 & 14.19 & 0.692 & \textbf{84.61} & 80.31 & \total{83.75}{-0.24} \\
 & \hl\textbf{+PreSTP} & \hl\textbf{5.21} & \hl\textbf{1.52$\times$} & \hl\textbf{15.5}
   & \hl\textbf{0.109} & \hl\textbf{20.64} & \hl\textbf{0.918}
   & \hl 84.59 & \hl\textbf{81.27} & \hl\total{\textbf{83.92}}{-0.07} \\
\midrule
\multirow{4}{*}{AnyFlow}
 & Vanilla   & 26.21 & 1.00$\times$ & 3.1 & --    & --    & --    & 85.63 & 80.88 & 84.68 \\
 & +TeaCache  & 24.44 & 1.07$\times$ & 3.3 & 0.354 & 14.58 & 0.759 & 85.32 & 80.21 & \total{84.30}{-0.38} \\
 & +FlowCache & 24.41 & 1.07$\times$ & 3.3 & 0.355 & 14.58 & 0.759 & 85.26 & 80.16 & \total{84.24}{-0.44} \\
 & \hl\textbf{+PreSTP} & \hl\textbf{8.19} & \hl\textbf{3.20$\times$} & \hl\textbf{9.9}
   & \hl\textbf{0.128} & \hl\textbf{20.49} & \hl\textbf{0.929}
   & \hl 85.33 & \hl\textbf{80.90} & \hl\total{\textbf{84.44}}{-0.24} \\
\bottomrule
\end{tabular}%
}
\caption{Comparison with relevant baselines on the 5-second video generation task. LPIPS, PSNR, and SSIM are also computed against the dense output. The best results among acceleration methods are in \textbf{bold}.}
\label{tab:main}
\vspace{-2mm}
\end{table*}

\begin{table*}[!t]
\centering
\small
\setlength{\tabcolsep}{5pt}
\resizebox{\textwidth}{!}{%
\begin{tabular}{ll ccc c cc ccc}
\toprule
\multirow{2}{*}{Model} & \multirow{2}{*}{Method}
 & \multirow{2}{*}{Latency(s)$\downarrow$}
 & \multirow{2}{*}{Speedup$\uparrow$}
 & \multirow{2}{*}{\makecell{Throughput\\(FPS)$\uparrow$}}
 & \multirow{2}{*}{$\Delta_{\text{sat}}\downarrow$}
 & \multicolumn{2}{c}{LPIPS$\downarrow$}
 & \multicolumn{3}{c}{VBench$\uparrow$} \\
\cmidrule(lr){7-8} \cmidrule(lr){9-11}
 & & & & & & Early & Late & Quality & Semantic & Total \\
\midrule
\multirow{4}{*}{CausVid}
 & Vanilla   & 60.17 & 1.00$\times$ & 8.3 & 0.097 & --    & --    & 82.16 & 76.76 & 81.08 \\
 & +TeaCache  & 48.32 & 1.25$\times$ & 10.4 & 0.095 & 0.308 & \textbf{0.428} & 82.06 & 77.44 & \totalgain{81.14}{+0.06} \\
 & +FlowCache & 48.22 & 1.25$\times$ & 10.4 & 0.096 & 0.434 & 0.524 & 82.01 & 76.97 & \total{81.00}{-0.08} \\
 & \hl\textbf{+PreSTP} & \hl\textbf{30.67} & \hl\textbf{1.96$\times$} & \hl\textbf{16.3}
   & \hl\textbf{0.051} & \hl\textbf{0.156} & \hl 0.530
   & \hl\textbf{82.18} & \hl\textbf{78.11} & \hl\totalgain{\textbf{81.36}}{+0.28} \\
\midrule
\multirow{4}{*}{Self-Forcing}
 & Vanilla   & 60.07 & 1.00$\times$ & 8.3 & 0.118 & --    & --    & 83.11 & 77.01 & 81.89 \\
 & +TeaCache  & 46.63 & 1.29$\times$ & 10.7 & 0.121 & 0.454 & 0.591 & 82.88 & 76.98 & \total{81.70}{-0.19} \\
 & +FlowCache & 48.80 & 1.23$\times$ & 10.3 & 0.119 & 0.451 & 0.587 & 82.95 & 77.11 & \total{81.78}{-0.11} \\
 & \hl\textbf{+PreSTP} & \hl\textbf{40.07} & \hl\textbf{1.50$\times$} & \hl\textbf{12.5}
   & \hl\textbf{0.102} & \hl\textbf{0.168} & \hl\textbf{0.560}
   & \hl\textbf{83.04} & \hl\textbf{78.21} & \hl\totalgain{\textbf{82.07}}{+0.18} \\
\midrule
\multirow{4}{*}{LongLive}
 & Vanilla   & 53.38 & 1.00$\times$ & 8.9 & 0.021 & --    & --    & 83.30 & 80.54 & 82.75 \\
 & +TeaCache  & 42.91 & 1.24$\times$ & 11.1 & 0.019 & 0.410 & 0.474 & \textbf{83.59} & 80.50 & \totalgain{82.97}{+0.22} \\
 & +FlowCache & 47.13 & 1.13$\times$ & 10.1 & 0.021 & 0.402 & 0.465 & 83.57 & 80.55 & \totalgain{82.97}{+0.22} \\
 & \hl\textbf{+PreSTP} & \hl\textbf{37.13} & \hl\textbf{1.44$\times$} & \hl\textbf{12.8}
   & \hl\textbf{0.007} & \hl\textbf{0.127} & \hl\textbf{0.309}
   & \hl 83.39 & \hl\textbf{80.95} & \hl\totalgain{82.90}{+0.15} \\
\bottomrule
\end{tabular}%
}
\caption{Comparison on the 30-second generation evaluated with VBench-Long metrics.
$\Delta_{\text{sat}}$ denotes the color saturation drift.}
\label{tab:main30s}
\vspace{-3mm}
\end{table*}

\paragraph{Content-adaptive mask reuse.}
However, obtaining $M_s^{(i)}$ requires at least one dense step, which is a non-negligible probing overhead in few-step generators. We exploit the temporal smoothness of video content: spatial redundancy changes gradually across adjacent frames, so
the pattern in frame $i{-}1$ is a reliable proxy for that in frame
$i$. This enables static-mask reuse without reconstruction at
every frame.

We use the cross-frame static-token ratio to gate static-mask reuse,
\begin{equation}
    \rho_i = \frac{\left|M_s^{(i)}\right|}{\left|\mathcal{I}^{(i)}\right|} .
\end{equation}
If $\rho_{i-1}\ge g$, frame $i$ inherits the preceding frame's static mask, $M_s^{(i)} = M_s^{(i-1)}$, instead of reconstructing it through dense probing; otherwise, $M_s^{(i)}$ is reconstructed by the cross-frame static masking procedure above. This gate applies to both multi-step and one-step models. Cross-step convergence contributes to token pruning within each frame but does not enter the reuse gate. To limit drift under prolonged reuse, dense probing is enforced every $\Delta$ frames to refresh the static mask.

This content-adaptive strategy reuses masks in static or slowly varying segments while reconstructing them for dynamic content, reducing probing overhead and limiting stale-mask error accumulation over long autoregressive rollouts. The complete inference procedure, including content-adaptive mask reuse, is summarized in Algorithm~\ref{alg:pstp}.

\newpage
\section{Experiments}
\subsection{Experimental Setup}

We evaluate on four distilled few-step AR video diffusion models. We adopt CausVid~\cite{yin2025slow} and Self-Forcing~\cite{huang2026self} as representative models for the common setting. In addition, we include the recent AnyFlow~\cite{gu2026anyflowanystepvideodiffusion} for the 5-second setting, and LongLive~\cite{yang2025longlive}, a native long-video model, for the 30-second setting. All models share the Wan2.1-1.3B backbone, use $4$ denoising steps, and generate at $832\times480$ resolution in a chunk-by-chunk manner. For short-video generation ($5$s, $81$ frames), we adopt VBench~\cite{huang2024vbench} to assess visual quality and semantic alignment on the official set of $946$ prompts, using five random seeds per prompt to reduce sampling variance. For the 30-second setting, we use VBench-Long~\cite{huang2025vbench++}. To measure fidelity to the base model, we further report LPIPS, PSNR, and SSIM against the original (dense) output, and quantify efficiency by diffusion latency and its speedup over the vanilla model. We compare against two representative training-free cache-based methods for AR video diffusion, TeaCache~\cite{liu2025timestep} and FlowCache~\cite{ma2026flow}. PreSTP usually adopts $\tau_s=0.97$, $\tau_f=0.997$, gate threshold $g=0.50$, re-anchor period $\Delta=3$, and a chunk size of 3 frames. Complete parameter setting configurations for PreSTP and both baselines are provided in the supplementary material.
\subsection{Main Results}

\begin{figure*}[!t]\centering
\includegraphics[width=\textwidth]{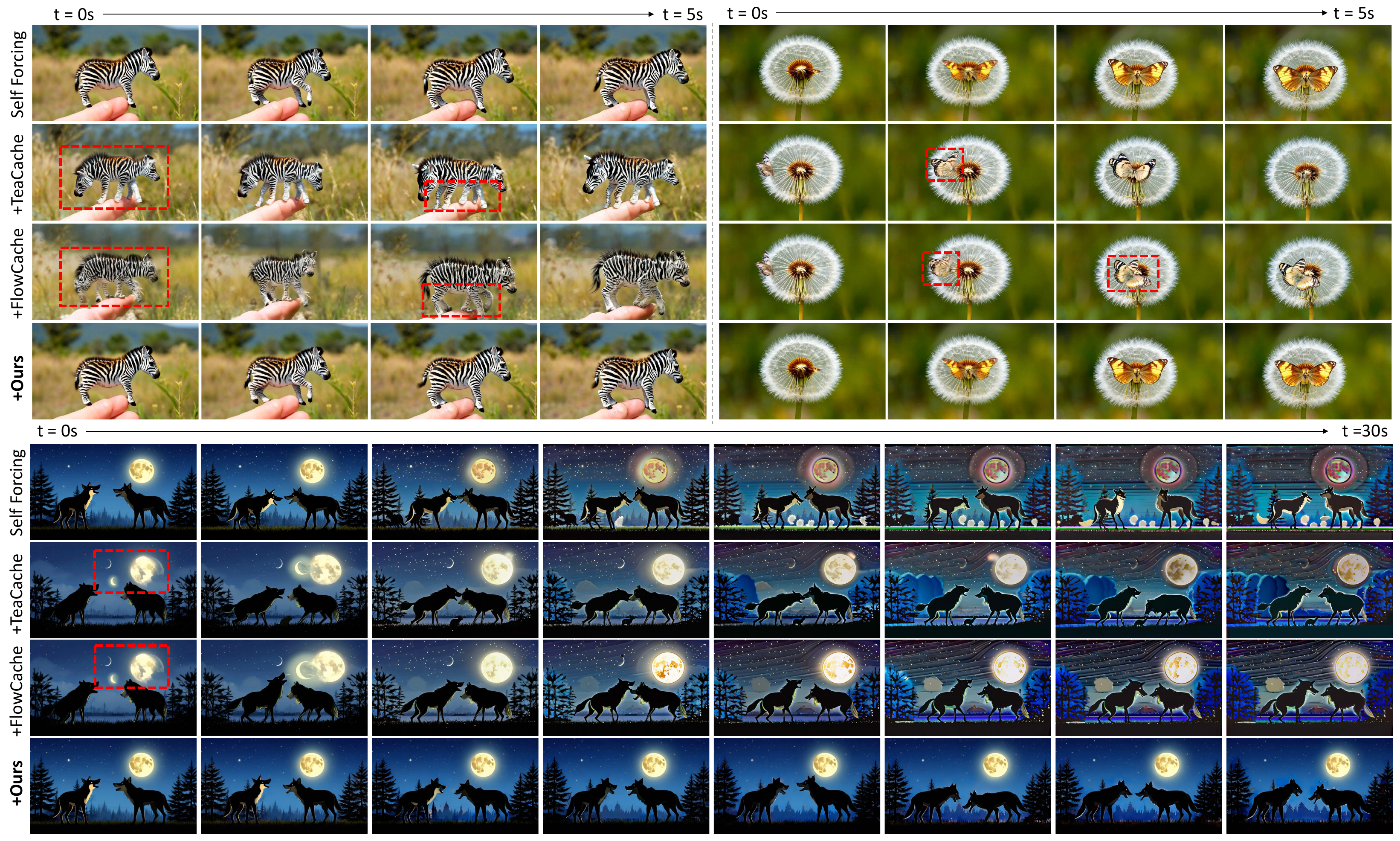}%

\caption{Qualitative comparison on the 5-second (top two) and 30-second (bottom) generation tasks. Red boxes highlight regions with visible degradation. More qualitative results and video comparisons are provided in the supplementary material.}
\vspace{-3mm}
\label{fig:qual}
\end{figure*}

\textbf{Short-video generation.}
Table~\ref{tab:main} reports the 5-second results. Across all three base models, our method attains the largest speedup, from $1.52\times$ to $3.20\times$, with the smallest fidelity loss relative to the dense output. The difference lies in the granularity of acceleration. TeaCache and FlowCache cache along the step axis, but distilled models retain little redundancy in denoising steps, so direct step reuse yields limited speedup at a clear cost in fidelity. Token-level pruning instead exploits the spatio-temporal redundancy that persists within each step, giving a substantially better acceleration-fidelity trade-off. On CausVid and Self-Forcing, it keeps VBench score within $0.07$ of the dense model and surpasses all baselines on LPIPS, PSNR, and SSIM by a wide margin. AnyFlow benefits the most. It re-encodes the KV cache of all historical frames rather than only the current denoising chunk, which raises its latency to $26.21$s and leaves heavy cross-frame redundancy. PreSTP removes this redundancy and reaches $3.20\times$ at only a $0.24$ drop in VBench score, whereas step-level caching stays near $1.07\times$ and degrades as its threshold grows, with TeaCache losing $2.48$ in VBench score for a mere $1.16\times$ at a threshold of $0.15$. Additional experimental settings and all 16 VBench dimensions are provided in the supplementary material.

\textbf{Long-video generation.}
Table~\ref{tab:main30s} reports the 30-second results on VBench-Long. A longer horizon accumulates more error, seen as artifacts and over-saturation, so we measure the saturation drift $\Delta_{\text{sat}}=|\text{sat}_{\text{late}}-\text{sat}_{\text{early}}|$ between the late and early phases as a proxy for long-horizon stability. Across all base models, PreSTP reaches the highest speedup, up to $1.96\times$, at the lowest $\Delta_{\text{sat}}$ and the smallest early-phase LPIPS. On CausVid and Self-Forcing, whose dense outputs exhibit substantial long-horizon drift, PreSTP improves VBench scores over dense generation. On Self-Forcing, for example, PreSTP achieves $82.07$, compared with $81.89$ for the dense model, $81.70$ for TeaCache, and $81.78$ for FlowCache. LongLive is a native long-video model with an already stable dense output, whose $\Delta_{\text{sat}}$ of $0.021$ is substantially lower than those of the other models. Even here PreSTP reaches the highest speedup, lowers $\Delta_{\text{sat}}$ to $0.007$, and raises the semantic score from $80.54$ to $80.95$. The small early-phase LPIPS shows PreSTP stays close to the original trajectory, while the higher Late LPIPS is less directly informative because the drifted base model no longer offers a reliable reference. Re-anchoring static tokens to earlier clean frames slows this error accumulation and keeps the long-horizon output stable.

\textbf{Qualitative comparison.}
Figure~\ref{fig:qual} compares our method against the vanilla output and caching baselines on prompts sampled from MovieGen~\cite{polyak2024movie}. In the zebra case, the baselines suffer severe structural corruption, including a duplicated head, blurred stripe texture, and an incorrect leg count of five. In the dandelion case, the baselines blur the butterfly in the early frames and distort its wing structure, whereas our method keeps the wings sharp and well-formed. This contrast shows that our pruning preserves both global structure and high-frequency detail, since our first-frame dense warm-up preserves the original denoising trajectory and keeps the output highly faithful to the dense reference. In the 30-second
case, the baselines accumulate temporal drift and produce spurious duplicated moons, whereas our method sustains strong subject and background consistency and remains even more stable than the vanilla output under long-horizon drift. This agrees with the lowest $\Delta_{\text{sat}}$ in Table~\ref{tab:main30s}, confirming PreSTP's long-horizon color stability.

\begin{table}[t]
\centering
\small
\setlength{\tabcolsep}{4pt}
\renewcommand{\arraystretch}{1.02}
\begin{tabular*}{0.72\textwidth}{@{\extracolsep{\fill}}llccccc@{}}
\toprule
Model & Method & Latency(s)$\downarrow$ & FPS$\uparrow$ & Speedup$\uparrow$ & LPIPS$\downarrow$ & VBench$\uparrow$ \\
\midrule
\multicolumn{7}{c}{\textit{5-second generation}} \\
\midrule
\multirow{2}{*}{OF}
 & Vanilla & 4.75 & 17.1 & 1.00$\times$ & -- & \total{84.32}{\phantom{-0.00}} \\
 & \textbf{+PreSTP} & \textbf{4.00} & \textbf{20.3} & \textbf{1.19$\times$} & \textbf{0.111} & \total{84.13}{-0.19} \\
\multirow{2}{*}{CF++}
 & Vanilla & 4.30 & 18.8 & 1.00$\times$ & -- & \total{83.40}{\phantom{-0.00}} \\
 & \textbf{+PreSTP} & \textbf{3.68} & \textbf{22.0} & \textbf{1.17$\times$} & \textbf{0.084} & \total{83.28}{-0.12} \\
\midrule
\multicolumn{7}{c}{\textit{30-second generation}} \\
\midrule
\multirow{2}{*}{OF}
 & Vanilla & 40.05 & 12.5 & 1.00$\times$ & -- & \total{79.56}{\phantom{+0.00}} \\
 & \textbf{+PreSTP} & \textbf{32.76} & \textbf{15.3} & \textbf{1.22$\times$} & \textbf{0.155} & \totalgain{79.72}{+0.16} \\
\multirow{2}{*}{CF++}
 & Vanilla & 33.07 & 15.2 & 1.00$\times$ & -- & \total{76.81}{\phantom{+0.00}} \\
 & \textbf{+PreSTP} & \textbf{30.04} & \textbf{16.7} & \textbf{1.10$\times$} & \textbf{0.109} & \totalgain{80.02}{+3.21} \\
\bottomrule
\end{tabular*}
\caption{Generalization on one-step AR models. LPIPS is computed against
the dense output (early phase for the 30-second setting). OF: One-Forcing; CF++: Causal Forcing++.}
\label{tab:onestep}
\end{table}

\subsection{Generalization to One-Step AR Models}
We further push to the most extreme setting, where the base model performs only a single denoising step. Here the denoising axis collapses and step-wise caching becomes inapplicable. We evaluate two representative one-step AR models, One-Forcing~\cite{feng2026oneforcingstableonestepautoregressive} and Causal Forcing++~\cite{zhao2026causalforcingscalablefewstep}, under both the 5-second and 30-second settings. As shown in Table~\ref{tab:onestep}, our method still delivers $1.10\times$ to $1.22\times$ speedups and higher throughput on every model, while preserving or improving VBench scores. Because it also prunes along the cross-frame axis, which does not depend on the number of denoising steps, PreSTP remains effective even when the step dimension collapses to a single step and step-wise caching no longer applies. For Causal Forcing++ at 30 seconds, where the dense model exhibits pronounced long-horizon degradation, re-anchoring static tokens to earlier clean frames even lifts the Total score from $76.81$ to $80.02$, with complete metrics in the supplementary material, confirming the effectiveness of cross-frame re-anchoring.

\subsection{Ablation Study}

\begin{figure}[H]
\centering
\begin{minipage}[t][0.33\textwidth][t]{0.55\textwidth}
\centering
\vspace{0pt}
\setlength{\tabcolsep}{2pt}
\renewcommand{\arraystretch}{1.35}
\resizebox{0.89\linewidth}{!}{%
\begin{tabular}{lccccc}
\toprule
Method & Speedup$\uparrow$ & VBench$\uparrow$ & LPIPS$\downarrow$ & PSNR$\uparrow$ & SSIM$\uparrow$ \\
\midrule
Vanilla                    & 1.00$\times$ & 83.99 & --    & --    & --    \\
+ Spatial$^{\dagger}$      & 1.41$\times$ & 79.44 & 0.399 & 14.41 & 0.717 \\
+ Warm-up                  & 1.42$\times$ & 83.94 & 0.103 & 21.04 & 0.925 \\
+ Temporal                 & 1.44$\times$ & 83.90 & 0.103 & 20.99 & 0.925 \\
\hl\textbf{+ Reuse (full)} & \hl 1.52$\times$ & \hl 83.92 & \hl 0.109 & \hl 20.64 & \hl 0.918 \\
\bottomrule
\end{tabular}%
}
\vfill
\captionof{table}{Key component ablation on Self-Forcing for 5-second generation. $^{\dagger}$ denotes cross-frame spatial pruning without the first-frame dense warm-up strategy.}
\label{tab:ablation}
\end{minipage}
\hfill
\begin{minipage}[t][0.33\textwidth][t]{0.42\textwidth}
\centering
\vspace{0pt}
\includegraphics[width=0.72\linewidth]{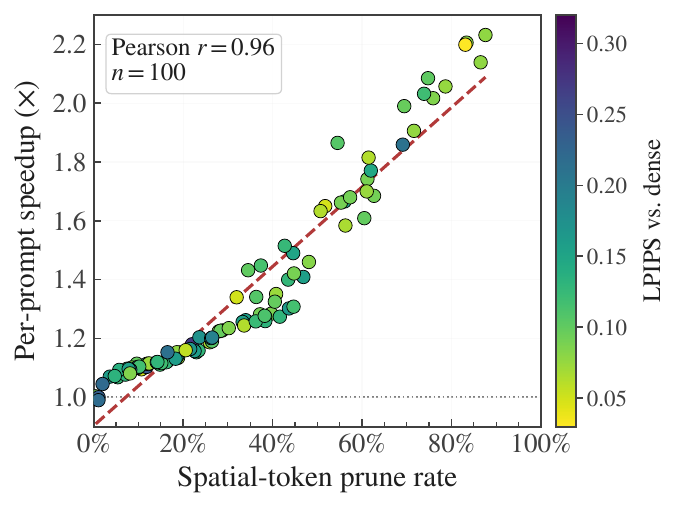}
\vfill
\captionof{figure}{\textbf{Content-adaptive acceleration.} Per-prompt speedup versus cross-frame pruning rate; color denotes LPIPS to dense output.}
\label{fig:adaptive}
\end{minipage}
\end{figure}

\textbf{Key component ablation.}
Table~\ref{tab:ablation} evaluates each component added
sequentially to the dense model on the 5-second task. Cross-frame
pruning without first-frame protection reaches $1.41\times$
speedup but severely degrades quality, reducing VBench to $79.44$.
Restoring the first-frame dense warm-up recovers VBench to $83.94$
at nearly the same speedup, confirming its importance for
trajectory fidelity at negligible cost. Cross-step pruning further
improves speedup at comparable fidelity, showing that the two masks
capture complementary token redundancy. Finally, content-adaptive mask reuse gives the full method its
highest speedup of $1.52\times$ at only a small drop ($-0.07$) in
VBench score.

\textbf{Content-adaptive acceleration.}
Figure~\ref{fig:adaptive} analyzes the first 100 MovieGen prompts. The online
cross-frame pruning rate ranges from $0$ to $88\%$, yielding per-prompt
speedups of $0.99\times$--$2.23\times$. Motion-intensive prompts remain close
to dense execution, whereas spatially redundant content permits larger gains.
The strong correlation between pruning rate and speedup ($r=0.96$) confirms
that computation adapts to online redundancy rather than a fixed token budget,
while LPIPS remains low across the observed range.

\begin{figure}[H]\centering
\captionsetup{width=\textwidth}
\includegraphics[width=\textwidth]{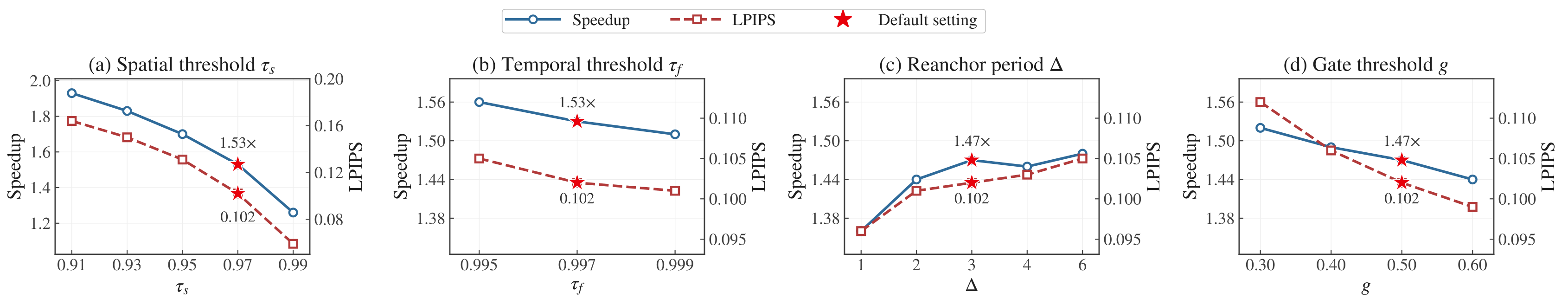}
\caption{Hyperparameter sensitivity on Self-Forcing for 5-second generation, conducted on a single seed for efficiency.}
\label{fig:hparam}
\vspace{-3mm}
\end{figure}

\textbf{Hyperparameter sensitivity.}
Figure~\ref{fig:hparam} varies each hyperparameter in turn with the others fixed. The
spatial threshold $\tau_s$ dominates the speedup-fidelity trade-off, as a lower value prunes
more cross-frame tokens and reaches up to $1.93\times$ at the cost of fidelity, while a
higher value stays closer to the dense trajectory. The remaining three, the temporal
threshold $\tau_f$, the reanchor period $\Delta$, and the gate threshold $g$, only shift the
two metrics within a narrow band. All curves change smoothly without sharp cliffs, so our
method is robust to the exact hyperparameter choice, and the default setting reaches a
$1.53\times$ speedup at an LPIPS of $0.102$.

\FloatBarrier

\section{Conclusion}

We presented PreSTP, a training-free framework for predictive
spatio-temporal token pruning in few-step AR video diffusion.
PreSTP uses clean estimates along the model's own denoising
trajectory to identify cross-frame static redundancy and
cross-step convergence, and combines them into a unified online
mask for skipping redundant token computation. Across diverse
few-step AR models, PreSTP achieves speedups of up to
$3.20\times$ on short-video generation and $1.96\times$ on
30-second generation while maintaining high fidelity to dense
outputs. It also improves long-horizon stability and remains
effective in one-step models, where step-wise caching is
inapplicable. These results demonstrate that token-level
spatio-temporal redundancy remains exploitable even after
few-step distillation has largely removed redundancy along the
denoising trajectory.

PreSTP's acceleration is content-dependent and becomes smaller
for motion-intensive videos with limited cross-frame redundancy.
In addition, it reduces query-side token computation while
attention still operates over the full historical KV context.
Future work will improve redundancy prediction for highly dynamic
content, jointly optimize token pruning and context compression,
and extend evaluation to broader video generation tasks such as
image-to-video synthesis.

\bibliographystyle{plainnat}
\bibliography{aaai2027}

\clearpage
\beginappendix
\setcounter{secnumdepth}{2}

This supplementary material provides implementation details and additional
experimental results supporting the main paper. Section~\ref{app:implementation}
describes the evaluation protocol and model-specific operating points.
Section~\ref{app:restoration} details last-computed-value restoration and
KV-cache consistency. Sections~\ref{app:vbench}--\ref{app:long_analysis}
report dimension-level VBench scores, extended one-step results,
hyperparameter analyses, and long-horizon stability. The final figures provide
additional qualitative comparisons. We use the notation defined in the main
paper.

\FloatBarrier

\section{Implementation and Evaluation Details}
\label{app:implementation}

\subsection{Evaluation Protocol}

We report diffusion latency, excluding prompt encoding, model loading, video
decoding, and metric evaluation. For generated videos indexed by $n$, global
throughput speedup is computed as
\begin{equation}
    \mathrm{Speedup}
    =
    \frac{\sum_n T^{\mathrm{dense}}_n}
         {\sum_n T^{\mathrm{method}}_n}.
\end{equation}
This dataset-level ratio, used throughout the main paper, differs from the mean
of per-video speedup ratios. For 5-second generation, we evaluate on the
official VBench set of 946 prompts using five random seeds per prompt to reduce
sampling variance. For 30-second generation, we follow the VBench-Long
protocol. LPIPS, PSNR, and SSIM measure fidelity to the dense output generated
from the same prompt and random seed.

\subsection{Method and Baseline Settings}

Table~\ref{tab:supp_method_settings} lists the PreSTP operating points used in
the reported experiments. CausVid and Self-Forcing use the common four-step
setting $(\tau_s,\tau_f,g,\Delta)=(0.97,0.997,0.50,3)$.
AnyFlow and LongLive use different similarity thresholds, as listed in
Table~\ref{tab:supp_method_settings}. All four-step experiments use chunks of
three frames. For one-step models, cross-step convergence is unavailable and
the corresponding mask is disabled. For both multi-step and one-step models,
the reuse gate uses the preceding frame's cross-frame static-token ratio,
with periodic dense probing to refresh the static mask.

\begin{table}[t]
    \centering
    \small
    \setlength{\tabcolsep}{2pt}
    \begin{tabular}{lcccccc}
        \toprule
        Model & Dur. & Steps & $\tau_s$ & $\tau_f$ & $g$ & $\Delta$ \\
        \midrule
        CausVid          & 5/30s & 4 & 0.97  & 0.997  & 0.50 & 3 \\
        Self-Forcing     & 5/30s & 4 & 0.97  & 0.997  & 0.50 & 3 \\
        AnyFlow          & 5 s    & 4 & 0.999 & 0.9995 & 0.40 & 2 \\
        LongLive         & 30 s   & 4 & 0.95  & 0.995  & default & default \\
        One-Forcing      & 5/30s & 1 & 0.985 & --     & 0.50 & 3 \\
        Causal Forcing++ & 5/30s & 1 & 0.98  & --     & 0.50 & 3 \\
        \bottomrule
    \end{tabular}
    \caption{Model-specific PreSTP operating points used in the reported
    experiments. A dash indicates that cross-step convergence is unavailable
    for one-step models; ``default'' denotes the native LongLive setting.}
    \label{tab:supp_method_settings}
\end{table}

CausVid and Self-Forcing restore skipped positions with their last-computed
clean values. For AnyFlow-FAR, the 5-second PreSTP configuration disables
historical KV-cache compression and follows the incremental cache-update
strategy of Self-Forcing. Each update encodes only the newly generated chunk
while retaining previously cached KV states, avoiding repeated encoding of
historical frames. LongLive uses its block-uniform
long-video variant. The first generated frame is computed densely for both
one-step models.

Table~\ref{tab:supp_baseline_settings} records the baseline thresholds that
define the operating points reported in the main paper. TeaCache and FlowCache
are applied without retraining. Their operating points are selected for each
model and duration because the scale of the cache criterion differs across
implementations.

\begin{table}[t]
    \centering
    \small
    \setlength{\tabcolsep}{3pt}
    \begin{tabular}{lccc}
        \toprule
        Base model & Duration & TeaCache $\tau$ & FlowCache $\tau$ \\
        \midrule
        Self-Forcing & 5 s  & 0.103 & 0.30 \\
        Self-Forcing & 30 s & 0.103 & 0.15 \\
        CausVid      & 5 s  & 0.08  & 0.40 \\
        CausVid      & 30 s & 0.08  & 0.15 \\
        AnyFlow      & 5 s  & 0.103 & 0.30 \\
        LongLive     & 30 s & 0.103 & 0.15 \\
        \bottomrule
    \end{tabular}
    \caption{Model- and duration-specific operating points of TeaCache and
    FlowCache used for the comparisons reported in the main paper.}
    \label{tab:supp_baseline_settings}
\end{table}

\begin{table*}[!t]
    \centering
    \small
    \setlength{\tabcolsep}{3pt}
    \begin{tabular}{llcccccccc}
        \toprule
        Model & Method & Subj. Cons. & Back. Cons. & Flicker & Motion
        & Dynamic & Aesthetic & Imaging & Quality \\
        \midrule
        \multirow{2}{*}{Self-Forcing}
        & Dense  & 95.74 & 96.35 & 99.13 & 98.37 & 64.72 & 65.81 & 69.16 & 84.67 \\
        & PreSTP & 95.70 & 96.45 & 99.32 & 98.37 & 62.50 & 65.71 & 69.23 & 84.59 \\
        \midrule
        \multirow{2}{*}{CausVid}
        & Dense  & 96.61 & 96.24 & 99.46 & 98.16 & 58.33 & 64.87 & 68.07 & 84.02 \\
        & PreSTP & 96.78 & 96.46 & 99.52 & 98.20 & 55.56 & 64.50 & 68.18 & 83.89 \\
        \bottomrule
    \end{tabular}
    \caption{Complete VBench quality-dimension scores for 5-second generation
    on Self-Forcing and CausVid.}
    \label{tab:supp_vbench_quality}
\end{table*}

\begin{table*}[!t]
    \centering
    \small
    \setlength{\tabcolsep}{2.5pt}
    \resizebox{\textwidth}{!}{%
    \begin{tabular}{llccccccccccc}
        \toprule
        Model & Method & Object & Multiple & Action & Color & Spatial & Scene
        & Appearance & Temp. Style & Overall & Semantic & Total \\
        \midrule
        \multirow{2}{*}{Self-Forcing}
        & Dense  & 94.68 & 87.56 & 96.40 & 87.17 & 82.55 & 57.75 & 20.49 & 24.46 & 26.89 & 81.26 & 83.99 \\
        & PreSTP & 94.56 & 87.10 & 96.60 & 87.77 & 82.22 & 57.82 & 20.47 & 24.50 & 26.90 & 81.27 & 83.92 \\
        \midrule
        \multirow{2}{*}{CausVid}
        & Dense  & 93.04 & 85.64 & 96.20 & 86.65 & 73.36 & 52.56 & 20.21 & 23.82 & 25.91 & 78.46 & 82.91 \\
        & PreSTP & 92.97 & 86.27 & 96.00 & 86.98 & 74.64 & 52.67 & 20.10 & 23.84 & 26.03 & 78.69 & 82.85 \\
        \bottomrule
    \end{tabular}%
    }
    \caption{Complete VBench semantic-dimension scores for 5-second generation
    on Self-Forcing and CausVid. ``Overall'' denotes overall consistency;
    Semantic and Total are aggregate scores.}
    \label{tab:supp_vbench_semantic}
\end{table*}

\begin{table*}[!t]
    \centering
    \small
    \setlength{\tabcolsep}{4pt}
    \begin{tabular}{llcccccccc}
        \toprule
        Model & Method & Latency & Speedup & Quality & Semantic & Total
        & LPIPS$_{\rm early}$ & LPIPS$_{\rm late}$ & $\Delta_{\rm sat}$ \\
        \midrule
        \multirow{2}{*}{One-Forcing}
        & Dense  & 40.05 & 1.00$\times$ & 81.79 & 70.63 & 79.56 & --    & --    & 0.085 \\
        & PreSTP & 32.76 & 1.22$\times$ & 81.38 & 73.08 & 79.72 & 0.155 & 0.498 & 0.064 \\
        \midrule
        \multirow{2}{*}{Causal Forcing++}
        & Dense  & 33.07 & 1.00$\times$ & 79.12 & 67.59 & 76.81 & --    & --    & 0.096 \\
        & PreSTP & 30.04 & 1.10$\times$ & 81.05 & 75.92 & 80.02 & 0.109 & 0.682 & 0.110 \\
        \bottomrule
    \end{tabular}
    \caption{Complete 30-second results on one-step AR video diffusion models.
    Latency is measured in seconds, VBench-Long scores use the $0$--$100$
    scale, and $\Delta_{\rm sat}$ measures long-horizon saturation drift.}
    \label{tab:supp_onestep_long}
\end{table*}

\section{Restoration and KV-Cache Update}
\label{app:restoration}

This section provides further details on the restoration and cache-consistency
rule summarized in the main paper. Given the keep set $S^{(i,k)}$, a pruned
forward pass executes the transformer only on selected tokens, while the
denoising solver still requires a complete token tensor after each step. For
every token, we maintain $l^{(i,k)}$, its most recent value produced by an
actual transformer computation. The complete output at step $k$ is
\begin{equation}
    [y^{(i,k)}]_j =
    \begin{cases}
        \big[\Phi_{\theta}(x_{t_k}^{(i)}, t_k, c,
        \mathcal{K}^{(<i)}; S^{(i,k)})\big]_j, & j \in S^{(i,k)},\\[2pt]
        [l^{(i,k-1)}]_j, & j \notin S^{(i,k)},
    \end{cases}
\end{equation}
and the last-computed buffer is updated only for tokens that are actually
computed,
\begin{equation}
    [l^{(i,k)}]_j =
    \begin{cases}
        [y^{(i,k)}]_j, & j \in S^{(i,k)},\\[2pt]
        [l^{(i,k-1)}]_j, & j \notin S^{(i,k)}.
    \end{cases}
\end{equation}
Thus, tokens skipped by either the cross-frame static mask or the cross-step
frozen mask follow the same restoration rule.

Each pruned transformer pass follows a gather--compute--scatter procedure.
Tokens in $S^{(i,k)}$ are gathered for computation, and their outputs and KV
states are written back to the original spatial positions. Outputs and cache
entries for skipped tokens remain unchanged.

\section{Full VBench Dimension Breakdown}
\label{app:vbench}

Tables~\ref{tab:supp_vbench_quality} and
\ref{tab:supp_vbench_semantic} report all 16 official VBench dimensions and
their aggregate scores for the two common four-step base models. All scores
are reported on the $0$--$100$ scale. PreSTP changes the Total score by only
$-0.07$ on Self-Forcing and $-0.06$ on CausVid, and most individual dimensions
remain close to the dense results. The most pronounced and consistent change
occurs in dynamic degree, which decreases from $64.72$ to $62.50$ on
Self-Forcing and from $58.33$ to $55.56$ on CausVid. This behavior is
consistent with static-token reuse slightly reducing measured motion, while
temporal consistency, appearance, and semantic alignment remain largely
preserved.

\section{Complete Results on One-Step Models}
\label{app:onestep}

Table~\ref{tab:supp_onestep_long} reports the complete 30-second metrics for
the one-step experiments reported in the main paper. Early and late LPIPS are
computed over the first and final quarters of each generated video,
respectively. PreSTP reduces latency on both models. On Causal Forcing++,
PreSTP increases the VBench-Long Total score from $76.81$ to $80.02$, with a
$1.10\times$ speedup.

\begin{table*}[!t]
\centering
\begin{minipage}[t]{0.48\textwidth}
    \centering
    \small
    \setlength{\tabcolsep}{2pt}
    \begin{tabular}{@{}llcccc@{}}
        \toprule
        Parameter & Value & Latency & Speedup & LPIPS & VBench \\
        \midrule
        \multirow{5}{*}{$\tau_s$}
        & 0.91 & 3.94 & 1.93$\times$ & 0.164 & 83.18 \\
        & 0.93 & 4.16 & 1.83$\times$ & 0.150 & 83.31 \\
        & 0.95 & 4.49 & 1.70$\times$ & 0.131 & 83.45 \\
        & \textbf{0.97} & 4.99 & 1.53$\times$ & 0.102 & 83.63 \\
        & 0.99 & 6.03 & 1.26$\times$ & 0.059 & 83.63 \\
        \midrule
        \multirow{3}{*}{$\tau_f$}
        & 0.995 & 4.90 & 1.56$\times$ & 0.105 & 83.70 \\
        & \textbf{0.997} & 4.99 & 1.53$\times$ & 0.102 & 83.63 \\
        & 0.999 & 5.05 & 1.51$\times$ & 0.101 & 83.57 \\
        \bottomrule
    \end{tabular}
    \captionof{table}{Threshold sensitivity on Self-Forcing for 5-second
    generation. Latency is measured in seconds, and all parameters other than
    the one being varied remain fixed at the selected setting.}
    \label{tab:supp_threshold_sensitivity}
\end{minipage}\hfill%
\begin{minipage}[t]{0.48\textwidth}
    \centering
    \small
    \setlength{\tabcolsep}{2pt}
    \begin{tabular}{@{}llcccc@{}}
        \toprule
        Parameter & Value & Latency & Speedup & LPIPS & VBench \\
        \midrule
        \multirow{5}{*}{$\Delta$}
        & 1 & 5.83 & 1.36$\times$ & 0.096 & 83.70 \\
        & 2 & 5.49 & 1.44$\times$ & 0.101 & 83.64 \\
        & \textbf{3} & 5.39 & 1.47$\times$ & 0.102 & 83.64 \\
        & 4 & 5.42 & 1.46$\times$ & 0.103 & 83.62 \\
        & 6 & 5.35 & 1.48$\times$ & 0.105 & 83.59 \\
        \midrule
        \multirow{4}{*}{$g$}
        & 0.30 & 5.23 & 1.52$\times$ & 0.112 & 83.45 \\
        & 0.40 & 5.31 & 1.49$\times$ & 0.106 & 83.58 \\
        & \textbf{0.50} & 5.39 & 1.47$\times$ & 0.102 & 83.64 \\
        & 0.60 & 5.50 & 1.44$\times$ & 0.099 & 83.70 \\
        \bottomrule
    \end{tabular}
    \captionof{table}{Mask-reuse sensitivity on Self-Forcing for 5-second
    generation, measured in a separate runtime environment with a dense-reference
    latency of $7.94$\,s. Speedups are relative to this dense reference.}
    \label{tab:supp_reuse_sensitivity}
\end{minipage}
\end{table*}

Over all frames, PreSTP obtains LPIPS, PSNR, and SSIM scores of $0.337$,
$13.75$, and $0.752$ on One-Forcing, and $0.481$, $11.54$, and $0.570$ on
Causal Forcing++. Since the dense reference can itself drift over long rollouts,
these fidelity scores should be considered alongside early LPIPS and
VBench-Long: the former measures agreement with the dense output in the first
quarter of the video, while the latter evaluates overall generation quality.

\section{Detailed Hyperparameter Analysis}
\label{app:hparams}

The main paper reports one-at-a-time hyperparameter sensitivity curves.
Tables~\ref{tab:supp_threshold_sensitivity} and \ref{tab:supp_reuse_sensitivity} provide
the corresponding numerical results. For efficiency, each setting is evaluated
using one seed per prompt on the official VBench set. The threshold experiments
in Table~\ref{tab:supp_threshold_sensitivity} and the mask-reuse experiments in
Table~\ref{tab:supp_reuse_sensitivity} use different runtime environments, with
dense-reference latencies of $7.62$\,s and $7.94$\,s, respectively. Each speedup
is computed against the dense reference in the same environment; absolute
latencies are not directly comparable across the two groups or with the main
results. The spatial threshold $\tau_s$ provides the primary control
over the speed--fidelity trade-off: lowering it marks more cross-frame tokens
as redundant and therefore increases both pruning and deviation from the dense
trajectory. Over the tested ranges, varying $\tau_f$, $g$, or $\Delta$ has a
smaller effect on speedup and LPIPS than varying $\tau_s$.

\section{Long-Horizon Stability}
\label{app:long_analysis}
We further analyze long-horizon redundancy, acceleration, and temporal drift
on Self-Forcing. Unlike paired fidelity metrics, the following diagnostics do
not assume that the dense trajectory remains stable throughout the rollout.

\begin{figure*}[!t]
    \centering
    \includegraphics[width=0.55\textwidth]{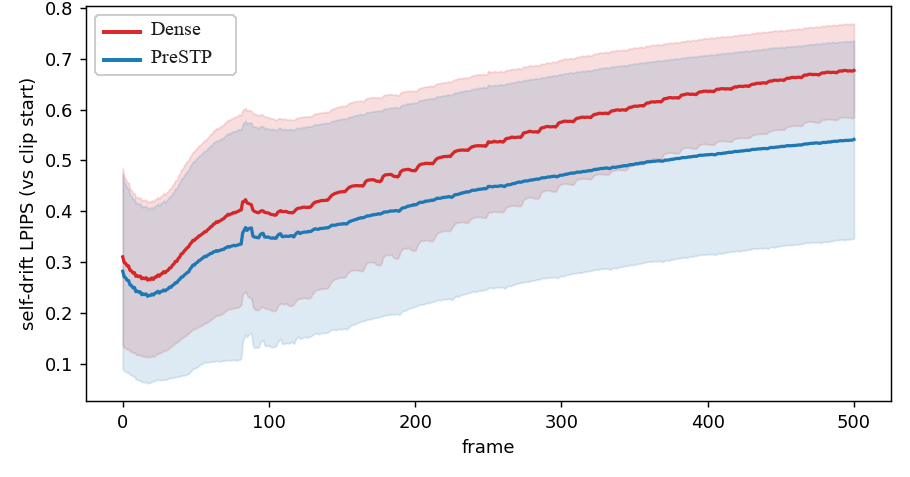}
    \caption{\textbf{Long-horizon self-drift on Self-Forcing.} LPIPS relative
    to the beginning of each video, averaged over the evaluation set. Each
    curve shows the mean with one standard deviation.}
    \label{fig:supp_self_drift}
\end{figure*}

\paragraph{Self-drift.}
Figure~\ref{fig:supp_self_drift} measures how far each generated frame departs
from the initial visual content as the autoregressive rollout progresses. Both
methods accumulate drift over time, but PreSTP remains consistently below the
dense model, with the gap becoming more pronounced later in the video. This
result indicates that reusing stable content does not amplify long-horizon
errors and can instead reduce their accumulation.

\begin{figure*}[!t]
    \centering
    \includegraphics[width=0.88\textwidth]{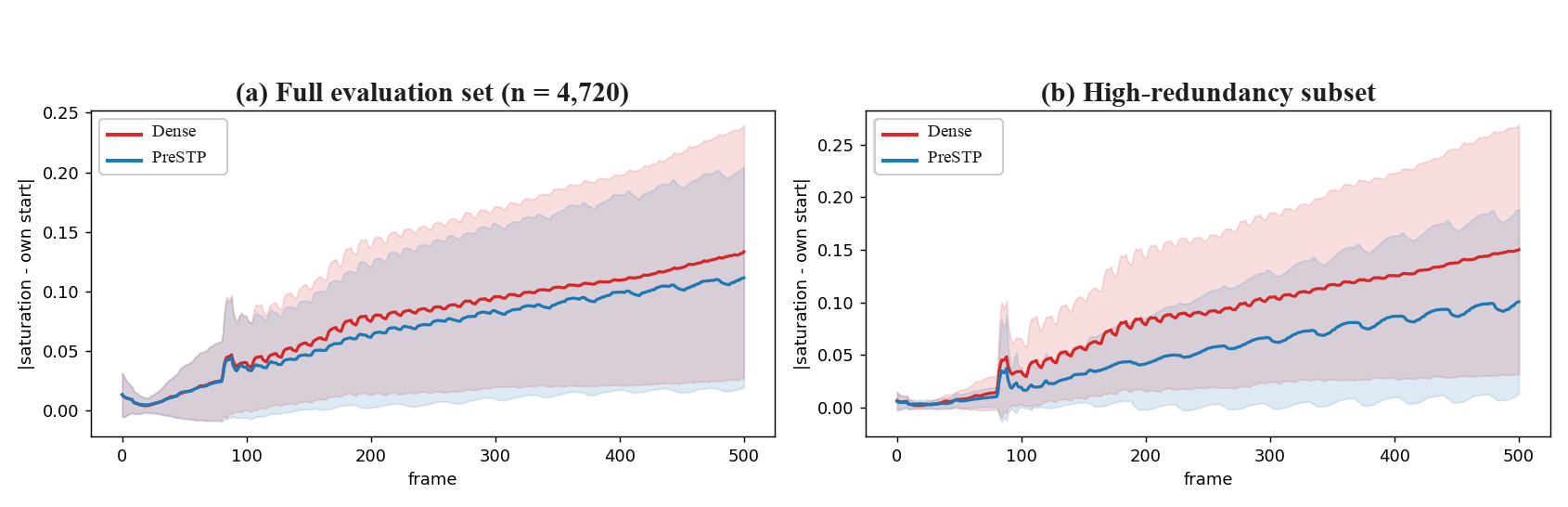}
    \caption{\textbf{Long-horizon saturation stability on Self-Forcing.}
    Saturation deviation from the beginning of each video over the complete
    evaluation set (left) and the high-redundancy subset (right). Each curve
    shows the mean with one standard deviation.}
    \label{fig:supp_saturation_drift}
\end{figure*}

\begin{figure*}[!t]
    \centering
    \includegraphics[width=\textwidth]{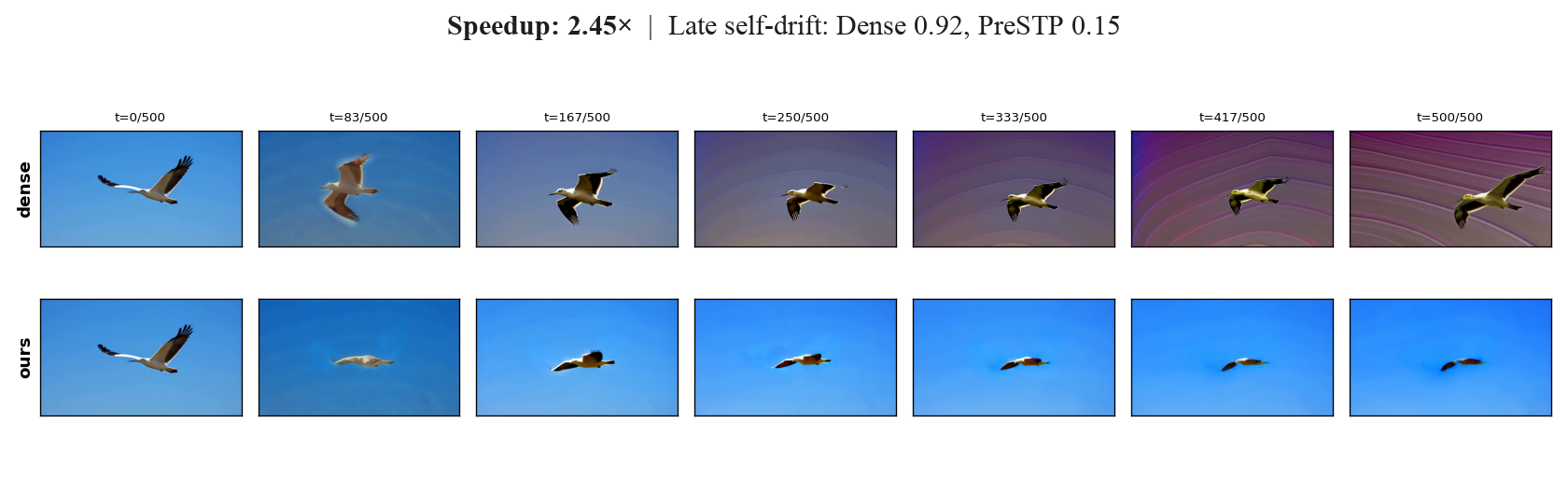}
    \caption{\textbf{Long-horizon qualitative comparison.} The largely static
    scene permits aggressive adaptive pruning. PreSTP preserves the blue-sky
    background and subject appearance, whereas the dense trajectory develops
    substantial color drift. Columns show corresponding rollout timestamps.}
    \label{fig:supp_long_qual}
\end{figure*}

\paragraph{Saturation drift.}
Figure~\ref{fig:supp_saturation_drift} separately evaluates long-horizon color
stability. PreSTP exhibits lower saturation deviation over the complete set,
and the reduction is substantially larger for videos with high spatial
redundancy. In this regime, periodic re-anchoring prevents repeatedly reused
static regions from inheriting the accumulated color shift of later dense
frames. This analysis complements VBench-Long by measuring temporal stability
without treating a potentially drifted dense output as the reference.

Figure~\ref{fig:supp_long_qual} illustrates the same behavior qualitatively.
Although the dense rollout progressively changes the background color and
appearance, PreSTP retains a stable blue sky and consistent subject throughout
the 30-second sequence. The example also shows why static content provides the
largest acceleration opportunity: its high cross-frame redundancy supports
aggressive pruning without sacrificing long-horizon consistency.

\section{Additional Qualitative Comparisons}
\label{app:qualitative}

\begin{figure*}[t!]
    \centering
    \includegraphics[width=0.90\textwidth]
        {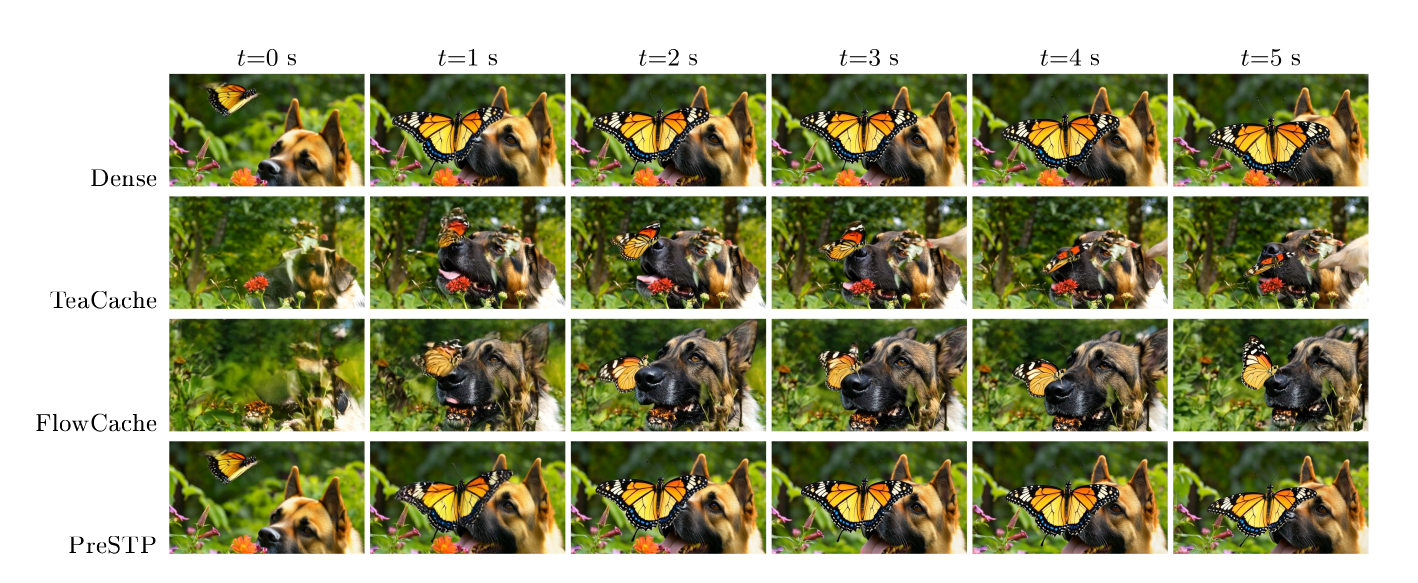}\\[3pt]
    \includegraphics[width=0.90\textwidth]
        {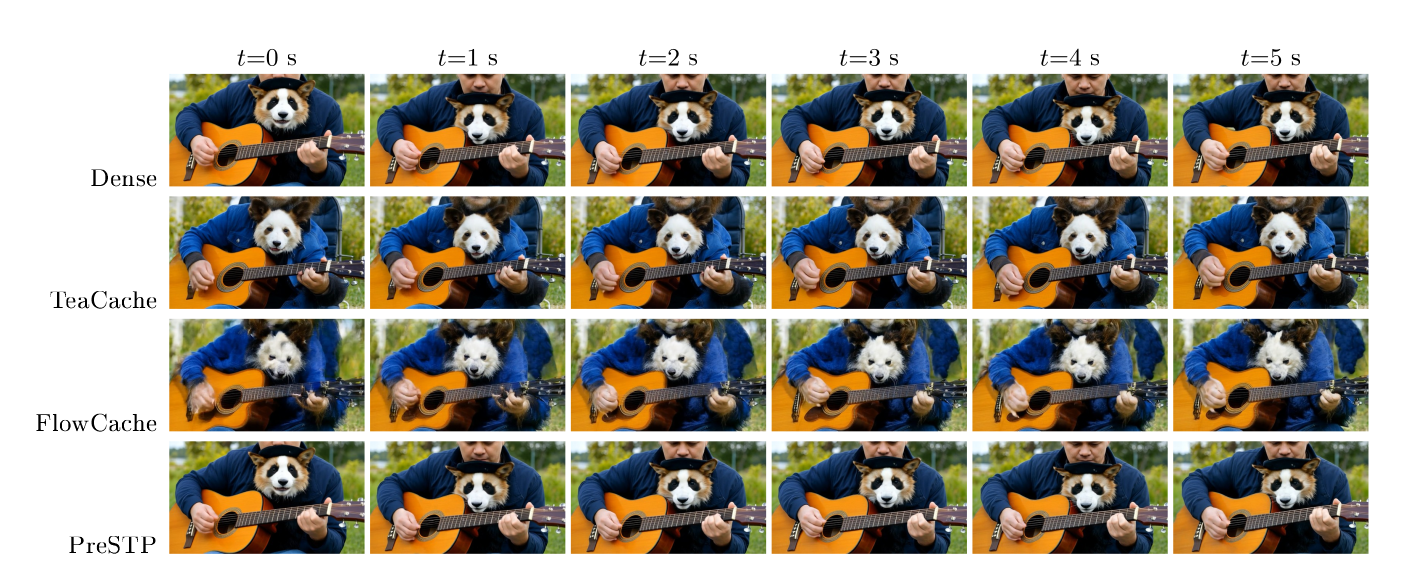}\\[3pt]
    \includegraphics[width=0.90\textwidth]
        {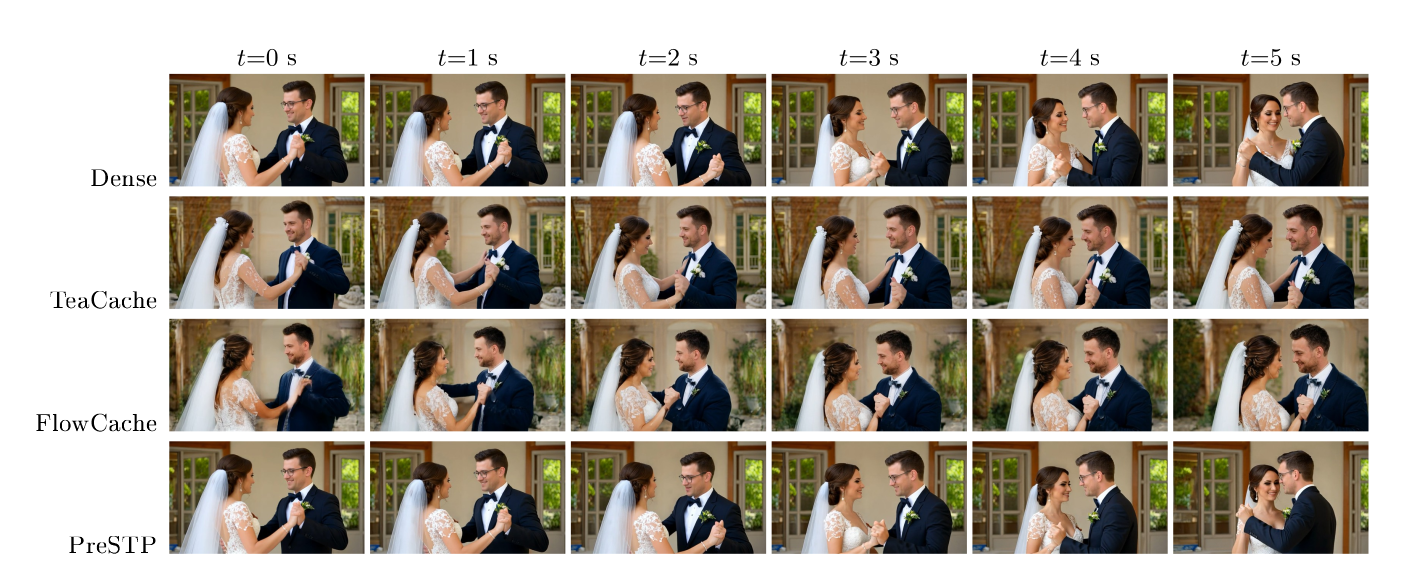}
    \caption{\textbf{Additional 5-second qualitative comparisons on
    Self-Forcing.} From top to bottom: a butterfly approaching a dog, a red panda
    sitting in a person's lap as the person plays a guitar, and a wedding interaction.}
    \label{fig:supp_qual_5s_a}
\end{figure*}

\begin{figure*}[t!]
    \centering
    \includegraphics[width=0.90\textwidth]
        {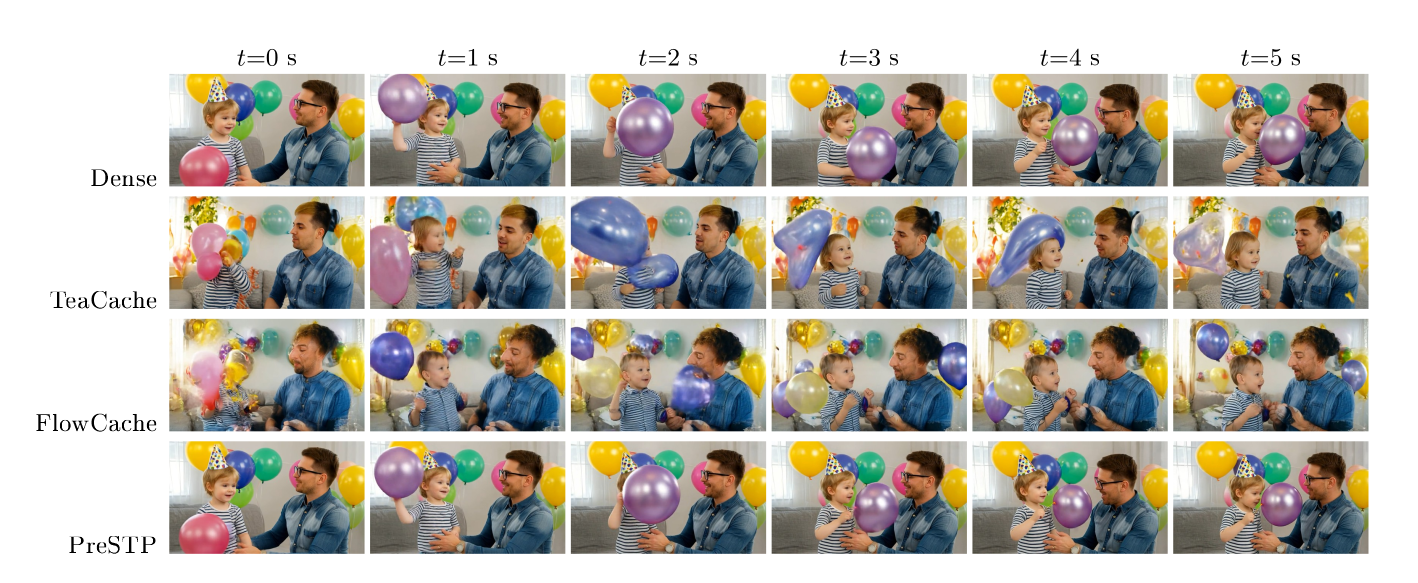}\\[3pt]
    \includegraphics[width=0.90\textwidth]
        {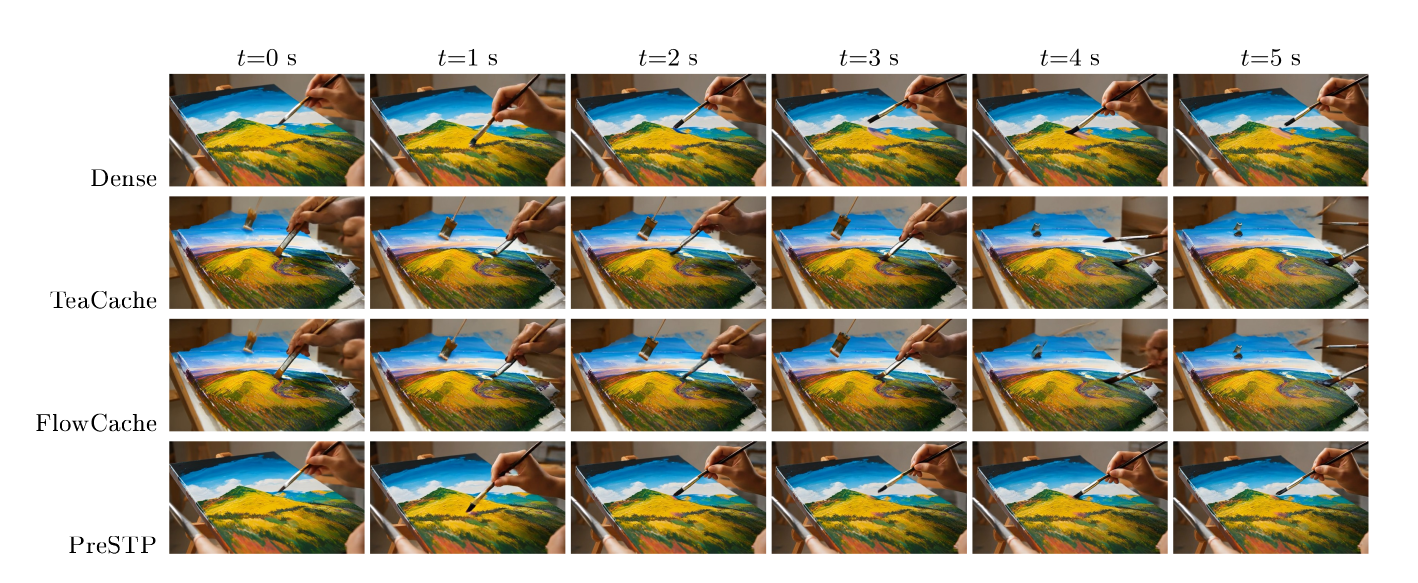}\\[3pt]
    \includegraphics[width=0.90\textwidth]
        {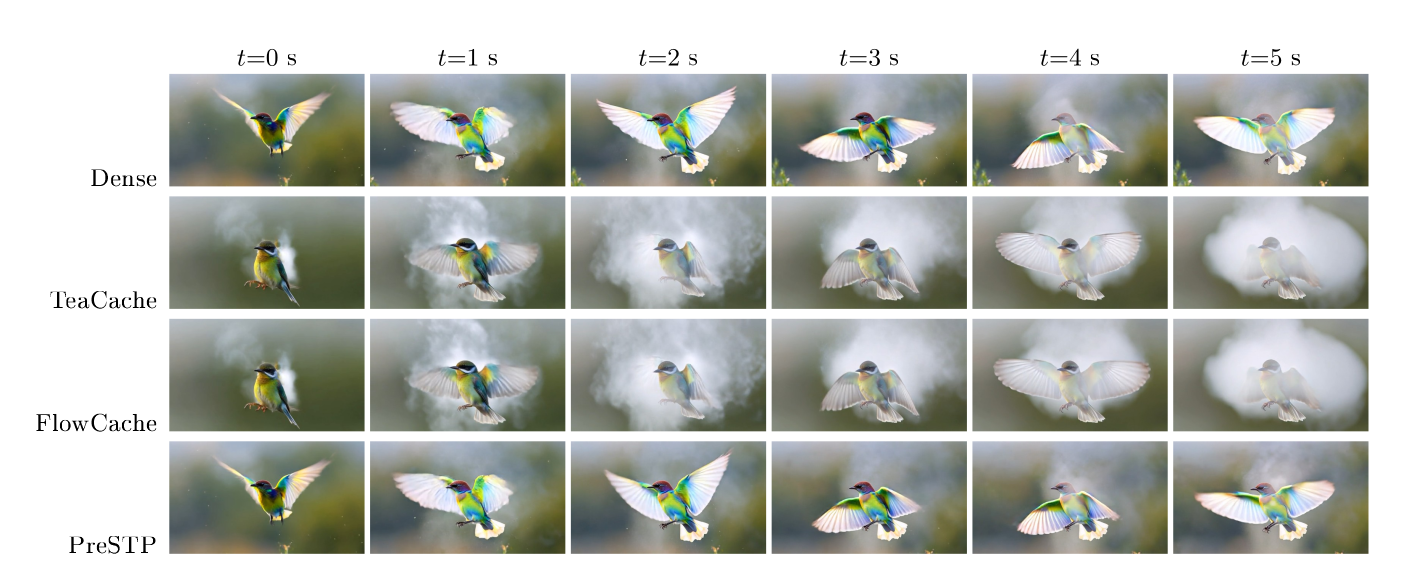}
    \caption{\textbf{Additional comparisons across diverse visual content.}
    From top to bottom: a parent and child playing with balloons, a person
    creating an oil painting with a brush, and a hummingbird in flight.}
    \label{fig:supp_qual_5s_b}
\end{figure*}

Figures~\ref{fig:supp_qual_5s_a} and \ref{fig:supp_qual_5s_b} compare dense
generation, TeaCache, FlowCache, and PreSTP on six additional 5-second examples
from Self-Forcing. Each comparison uses matched prompts and sampling
conditions across methods.

In Figure~\ref{fig:supp_qual_5s_a}, the caching baselines introduce visible
structural artifacts, including distortions of the dog's face and deformed
facial features of both the red panda and the person holding it. The guitar
scene also exhibits blurred hand structure, while the wedding sequence shows
altered human poses. In the displayed frames, PreSTP better preserves facial
structure and hand details.

In Figure~\ref{fig:supp_qual_5s_b}, TeaCache and FlowCache exhibit distorted
human heads in the balloon scene, malformed paintbrushes in the oil-painting
scene, and blurred or distorted hummingbird wings. These artifacts affect
the structure of people and objects, beyond changes in scene composition.
PreSTP shows fewer such distortions and preserves clearer local details in
the displayed frames.

\end{document}